\documentclass[english]{article}
\usepackage[T1]{fontenc}
\usepackage[latin9]{inputenc}
\usepackage{geometry}
\usepackage{float}
\usepackage{amstext}
\usepackage{amsthm}
\usepackage{amssymb}
\usepackage{graphicx}

\makeatletter

\floatstyle{ruled}
\newfloat{algorithm}{tbp}{loa}
\providecommand{\algorithmname}{Algorithm}
\floatname{algorithm}{\protect\algorithmname}

  \theoremstyle{plain}
  \newtheorem*{thm*}{\protect\theoremname}

\date{}

\makeatother

\usepackage{babel}
  \providecommand{\theoremname}{Theorem}

\begin{document}

\title{Toward a Better Understanding of Leaderboard}

\author{ZHENG Wenjie}
\maketitle
\begin{abstract}
The leaderboard in machine learning competitions is a tool to show
the performance of various participants and to compare them. However,
the leaderboard quickly becomes no longer accurate, due to hack or
overfitting. This article gives two pieces of advice to prevent easy
hack or overfitting. By following these advice, we reach the conclusion
that something like the Ladder leaderboard introduced in \cite{blum2015ladder}
is inevitable. With this understanding, we naturally simplify Ladder
by eliminating its redundant computation and explain how to choose
the parameter and interpret it. We also prove that the sample complexity
is cubic to the desired precision of the leaderboard.
\end{abstract}

\section{Introduction}

Machine learning competitions have been a popular platform for young
students to practice their knowledge, for scientists to apply their
expertise, and for industries to solve their data mining problems.
For instance, the Internet streaming media company Netflix held the
Netflix Prize competition in 2006, to find a better program to predict
user preferences. Kaggle, an online platform, hosts regularly competitions
since 2010.

These competitions are usually prediction problems. The participant
is given the independent variable $X$, and then he is required to
predict the dependent variable $Y$. Usually, the host divides his
data into three data sets: \emph{training}, \emph{validation} and
\emph{test}. The training dataset is fully available: every participant
(having a competition account) can download it and observe its $Y$
as well as its $X$. This allows them to build their models. The validation
data set is partially available to participants: they can only observe
its $X$. This dataset is used to construct the so-called leaderboard.
The participants submit their prediction of $Y$ to the host, and
the host calculates their scores and ranks, and show them on the leaderboard,
so that every participant could know his chance to win the competition.
The test dataset is private. They are only used once at the final
day to determine who is the final winner. Usually, the winner gets
a reward.

The reason that the final result is determined by the reserved test
set instead of the validation set is because the validation set could
be hacked. Since the participant could submit his prediction over
and over during the life of the competition, he has much chance to
improve his model's performance on the validation set, either by overfitting
or hacking. In consequence, by the final day, the score he gets on
the validation set may have been much higher than his model deserves.
This is why it is frequently observed that the final winner of the
competition is \emph{not} the ``winner'' on the leaderboard. 

Although the leaderboard has no effect on the decision of the final
winner, it could be quite annoying if it cannot truly reflect the
performance of each participant. Firstly, such a leaderboard allures
inexperienced participants to overfit the validation set. Secondly,
it encourages certain participants to hack the validation set in order
to get a fake temporary honor or to disturb the order of the competition.
Thirdly, it is not a good experience to see one's non-overfitting
model rank below someone hacking the validation set. It could be said
that during the whole life of the competition, the participants compete
around the leaderboard.

In view of this, some researchers tried to build an accurate leaderboard
by preserving the accuracy of the estimator of the loss function.
This could be hard since the participant can modify their model adaptively
according to the feedback they get from the leaderboard. \cite{hardt2014preventing,steinke2014interactive}
suggest that maintaining accurate estimates on a sequence of many
adaptively chosen classifiers may be computationally intractable.
In light of this, \cite{blum2015ladder} proposes the \emph{Ladder}
mechanism to restrain the feedback that the participant could get
from the leaderboard. The idea is that the participant gets a score
if and only if this score is higher than the best among the past by
some margin. By restraining the feedback, the participants have less
information to adapt their models, and thus less chance to hack the
leaderboard by overfitting the validation set.

However, \cite{blum2015ladder} fails to point out whether this kind
of mechanism is necessary: does there exist any other mechanism that
achieves the same or better effect? If we have to use Ladder, what
is its strength and shortcoming? How well can we hack the leaderboard?
Is is true that the first leader is better than the second? If so,
then by how much? There are two parameters in Ladder: margin $\eta$
and precision $\eta$; what is their relationship? Does the heuristic
way to choose $\eta$ provided in the paper have any theoretic guarantee?
If the participant holds many accounts, is Ladder still effective?

In our paper, we answer these questions. A better understanding of
the leaderboard will be achieved during the reading of this article.
First, we show that traditional leaderboard is easy to hack. In consequence,
something like Ladder mechanism is necessary. Then, we perform another
hack to show that a leaderboard cannot be arbitrarily accurate (Section
\ref{sec:Leaderboard-failure}). Afterwards, we recognize the essence
inside the Ladder mechanism and thus simplify it (Section \ref{sec:Simplified-Ladder-leaderboard}).
Finally, we generalize the Theorem 3.1 in \cite{blum2015ladder} to
take into consideration of the fact that each participant may be allowed
to possess multiple accounts. And we slightly improve the upper bound
as well (by eliminating the dependence on $n$ in the logarithmic
factor). Our result shows that, while using Ladder mechanism, the
leaderboard needs $\tilde{O}(M\epsilon^{-3})$ samples to control
the error within $\epsilon$, where $M$ is the number of accounts.\footnote{$\tilde{O}()$ stands for omitting the logarithm term.}
This result suggests that Ladder is relatively robust to number of
submissions, but may still be vulnerable to number of accounts (Section
\ref{sec:Sample-complexity}). In this article, we study the binary
classification competition, but our result can be straightforwardly
generalized to other kinds of competitions.

The highlight of this article is that we are not advertising any ``magic''
algorithm; we are just pursuing a better understanding of the leaderboard.
We depart from some basic property (robust against hacks and overfittings)
that a leaderboard should satisfy in order to protect its accuracy
and fairness. We reach the conclusion that something like Ladder is
inevitable. This understanding further allows us to eliminate the
redundant computation in the origin Ladder, and to only retain its
essence. We also give an upper bound. But we do not stop there. We
interpret this upper bound and use it as a tool to understand the
advantages as well as limitation of Ladder when used in practice.

\section{Leaderboard failure\label{sec:Leaderboard-failure}}

In this sections, we show some examples where the leaderboard is hackable
if it releases certain information. With these examples, we know at
least what to avoid when building a leaderboard.

\subsection{Full-information leaderboard is hackable}

In this subsection, we show that if a leaderboard shows the score
of each submission, this leaderboard is easy to hack.

Supposing that the validation set contains $n$ different data points
$S=\{(x_{1},y_{1}),(x_{2},y_{2}),\ldots,(x_{n},y_{n})\}$, where $y_{i}\in\{0,1\}$
for each $i$. The participant is expected to build a function $f=f(x)$,
so that $f(x)$ is a good estimator of $y$. Let the score be the
accuracy of this estimator, which is defined as $\textrm{score}(f)=\frac{1}{n}\sum_{i=1}^{n}1_{\{y_{i}=f(x_{i})\}}$
on the validation set. Every time the participant submits his $\hat{y_{i}}=f(x_{i})$
to the host, the host shows the score of $f$ to the participant via
the leaderboard. We show that this kind of leaderboard is easy to
hack.

To hack this leaderboard, we perform a boosting attack.\footnote{Actually this is not really a boosting technique, but we use the same
terminology as in \cite{blum2015ladder} here. } The idea is that if we have many independent submissions, whose accuracy
are only a little higher than 0.5, then we can combine them via majority
vote policy to construct a submission whose accuracy is much higher
than 0.5. 

In detail, we randomly pick a vector $u\in\{0,1\}^{n}.$ If its accuracy
is higher than 0.5, then we keep $v=u$, and otherwise $v=1_{n}-u$.
Having got $m$ such vectors $v^{1},v^{2},\ldots,v^{m}$, we construct
the submission $\hat{y}^{m}=(\hat{y}_{1}^{m},\hat{y}_{2}^{m},\ldots,\hat{y}_{n}^{m})^{T}$,
where $\hat{y}_{i}^{m}$ equals to 1, if $\frac{1}{m}\sum_{j=1}^{m}v_{i}^{j}>0.5$,
and equals to 0 otherwise. 

\begin{figure}
\begin{centering}
\includegraphics[scale=0.6]{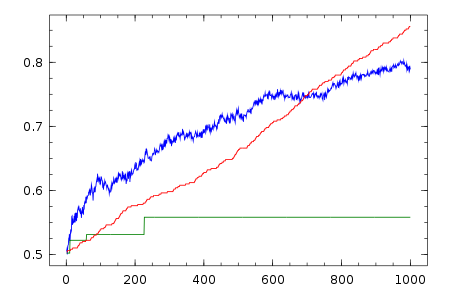}
\par\end{centering}
\caption{Attack on various leaderboard (sample size: 1000). Blue: boosting
attack on traditional leaderboard. Green: boosting attack on Ladder.
Red: brute-force enumeration attack on parameter-free Ladder. Blue
curve can also be seen as if a hacker uses boosting with 1000 accounts
to hack Ladder leaderboard.\label{fig:Attack}}
\end{figure}

Figure~\ref{fig:Attack} shows the result of this attack on a leaderboard
of 1000 samples. We see that within $10^{3}$ submissions, the hacker's
score climbs from 0.5 to 0.8 on the leaderboard. Therefore, in order
to protect the leaderboard from the boosting attack, we cannot release
information each time there is a submission. In consequence, we adopt
the idea that the leaderboard gives the participant feedback only
when his score is higher than the highest in the past.

\subsection{High-precision leaderboard is hackable}

Normally, a leaderboard shows two things \textendash{} score and rank.
In this subsection, we show that if a leaderboard precisely reflects
the ranks, then this leaderboard is easy to hack.

For this, we consider a minimal leaderboard, which shows nothing other
than the ranks. In other words, the participants are not able to observe
the scores. Furthermore, inheriting the argument from the previous
subsection, the leaderboard uses the highest score that a participant
has ever achieved to compute the rank. In other words, a participant
knows nothing even if he beats his old scores unless his new score
is higher enough to beat \emph{another} participant, whose rank was
higher than his. 

This leaderboard displays really little information. However, even
with so little information displayed, it is still hackable. For this,
we perform a brute-force enumeration attack. Precisely, the hacker
signs up two accounts A and B. At first, he uses A to submit a random
guess $u^{1}$, and he gets a rank $a_{1}$ for A. Then, he flips
one component in this submission (say, $u_{1}^{1}$ from 0 to 1, or
from 1 to 0), and uses B to submit it as $u^{2}$. He will get a rank
$b_{1}$ for B, which is different than $a_{1}$. Let us assume that
$b_{1}$ is higher than $a_{1}$(otherwise, we just switch the name
of account A and B). Then, he flips an unchanged component (say $u_{2}^{2}$)
and switches to A to submit it as $u^{3}$. The score $u^{3}$ yields
is either higher or lower than $u^{2}$. If it is lower than $u^{2}$
(must be equal to $u^{1}$ in this case), then he sees no change on
A's rank. In this case, he repeats this step by flipping another component
(say $u_{3}^{2}$) until it is higher than $u^{2}$ or all components
have been flipped once. If it is higher than $u^{2}$, since the leaderboard
precisely reflects the rank, it must move A's rank from $a_{1}$to
$a_{2}$, which is higher than $b_{1}$. Once again, the hacker switches
to the account B and repeats the process \ldots{} He gets $a_{1}\prec b_{1}\prec a_{2}\prec b_{2}\prec a_{3}\prec\cdots$.
Since the score is bounded by 1, and each score increment is constant,
the hacker finally gets the score 1, which also means getting all
answers right and ranking the highest on the leaderboard.

With this simple attack, we show that as long as the leaderboard precisely
reflects the rank, a hacker equipped with two accounts can achieve
arbitrarily high score. Therefore, a leaderboard should never reveal
precise ranks. In other words, there are cases where two participants
with different scores (this difference will not be observable) see
them ranked together on the leaderboard. 

\section{Simplified Ladder leaderboard\label{sec:Simplified-Ladder-leaderboard}}

In the previous section, we learned that a leaderboard should avoid
some pitfalls. In this section, we show that the Ladder leaderboard
\cite{blum2015ladder} successfully avoids them. We first introduce
Ladder and simplify it, and then demonstrate its robustness against
boosting and enumeration attacks. Throughout this paper, we use the
following notation.

\paragraph*{Notation.}

$[x]_{\eta}$ denotes the number $x$ rounded to the nearest integer
multiple of $\eta$; $\lfloor x\rfloor_{\eta}$ to the nearest not
higher than $x$; $\lceil x\rceil_{\eta}$ to the nearest not lower
than $x$. $\eta$ is a number in $(0,1]$, and is usually among the
values 0.1, 0.01, etc.. When $\eta$ is missing, it is consider to
be 1 in convention. $\log x$ denotes the binary logarithm.

The idea of Ladder is simple. The leaderboard only shows the best
score that a participant has ever achieved, and updates it only when
a record-breaking score is higher than it by some margin $\eta$ (Algorithm~\ref{alg:Original-Ladder}).
Notice that there is a parameter $\eta$ in this algorithm. To overcome
this inconvenience, \cite{blum2015ladder} also suggests a parameter-free
Ladder leaderboard. However, this parameter-free version is hackable
(Figure~\ref{fig:Attack}). The method is to make a first submission
containing half 1 and half 0. Then switch the place of a 1 and a 0
in each submission.

In this paper, we give a simplified Ladder version (Algorithm~\ref{alg:Simplified-Ladder})
as well as a way to choose the optimal value of $\eta$. Comparing
these two algorithms, the only tiny difference is the \emph{condition}
in Step~3 \textendash{} we drop the margin $\eta$. This raises naturally
the question whether this modification breaks the algorithm? No. In
fact, the margin $\eta$ is already captured in the assignment $R_{t}\leftarrow\left[h_{t}\right]_{\eta}$
by he precision $\eta$. Since $R_{t-1}$ is always an integer multiple
of $\eta$, $R_{t}$ can be greater than $R_{t-1}$ if and only if
$h_{t}$ is higher than $R_{t-1}$ by a margin of $\frac{\eta}{2}$.
The simplification does not stop here. Indeed, Step~3 can be rephrased
as $R_{t}=\max\left(R_{t-1},\left[h_{t}\right]_{\eta}\right)$. So
the idea of the simplified Ladder leaderboard is just to \emph{authentically}
display the best score achieved by each participant so far, but with
a certain level of \emph{precision} $\eta$. 

\begin{algorithm}[h]
\caption{Original Ladder \cite{blum2015ladder}\label{alg:Original-Ladder}}

Assign initial score $R_{0}\leftarrow-\infty$.

\textbf{for} round $t=1,2,\ldots$ \textbf{do}

\qquad{}1. Receive submission $u^{t}$

\qquad{}2. $h_{t}\leftarrow$score of $u^{t}$

\qquad{}3. \textbf{If} $h_{t}>R_{t-1}+\eta$ \textbf{then }$R_{t}\leftarrow\left[h_{t}\right]_{\eta}$
\textbf{else} $R_{t}\leftarrow R_{t-1}$

\qquad{}4. Show $R_{t}$ on the leaderboard

\textbf{end for}

Note: $\left[x\right]_{\eta}$ denotes the number $x$ rounded to
the nearest integer multiple of $\eta$.
\end{algorithm}

\begin{algorithm}[h]
\caption{Simplified Ladder\label{alg:Simplified-Ladder}}

Assign initial score $R_{0}\leftarrow-\infty$.

\textbf{for} round $t=1,2,\ldots$ \textbf{do}

\qquad{}1. Receive submission $u^{t}$

\qquad{}2. $h_{t}\leftarrow$score of $u^{t}$

\qquad{}3. \textbf{If} $h_{t}>R_{t-1}$ \textbf{then }$R_{t}\leftarrow\left[h_{t}\right]_{\eta}$
\textbf{else} $R_{t}\leftarrow R_{t-1}$

\qquad{}4. Show $R_{t}$ on the leaderboard

\textbf{end for}

Note: $\left[x\right]_{\eta}$ denotes the number $x$ rounded to
the nearest integer multiple of $\eta$.
\end{algorithm}

This understanding is very helpful. On the one hand, it simplifies
the implementation of the Ladder leaderboard. The practitioners have
less chance to make an error (e.g., by accidentally dropping the precision
$\eta$ or configure a smaller one). On the other hand, it greatly
simplifies the analysis. $\eta$ here is no longer an algorithmic
parameter. It is instead the precision which the leaderboard offers.

After the presentation of Ladder algorithm, now let us try to answer
this question: does Ladder avoid the pitfalls mentioned above? Yes.
On the one hand, it only reveals the highest score so far. On the
other hand, it does not give arbitrarily precise information \textendash{}
a participant yielding 0.644 is not distinguishable from another participant
yielding 0.636 when $\eta=0.01$. 

But there still remain questions that whether Ladder is hackable.
Is it robust against \emph{all} attacks besides the above mentioned
ones? The answer is \emph{yes}, provided that the attacker does not
possess \emph{many} accounts. And the robustness is proportional to
the cubic root of the size of the validation set. If we want the precision
$\eta$ to be $0.1$, we should have $10^{3}$ samples; if $\eta=0.01$,
we will need $10^{6}$ samples. This will be proved in the next section. 

Here again our understanding of Ladder contributes. If $\eta$ is
too large, which means that the precision is low, the participants
will not be distinguishable \textendash{} they are clustered on the
leaderboard. Such a leaderboard is not informative. If $\eta$ is
too small, which means the precision is too high and the leaderboard
reveals too much information, there will not be enough samples to
maintain the authenticity of the leaderboard \textendash{} the leaderboard
is easy to hack or overfit. Such a leaderboard is false. This trade-off
is the central topic of the next section.

\begin{figure}[H]
\begin{minipage}[t]{0.5\columnwidth}%
\begin{center}
\includegraphics[scale=0.5]{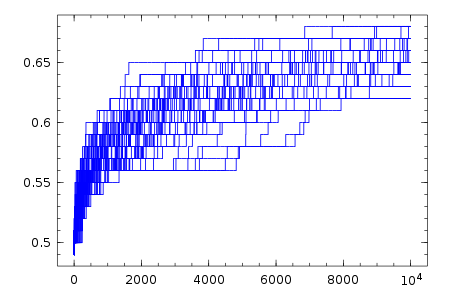}
\par\end{center}%
\end{minipage}%
\begin{minipage}[t]{0.5\columnwidth}%
\begin{center}
\includegraphics[scale=0.5]{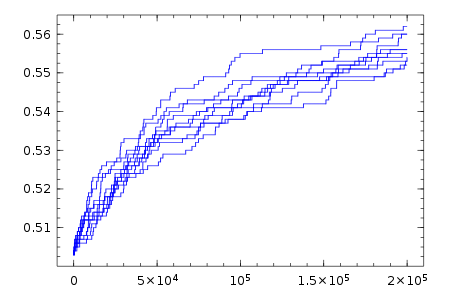}
\par\end{center}%
\end{minipage}

\caption{Attack on Ladder. Each trajectory is correspondent to an attack. Left:
n=1000, $\eta$=0.01. Right: n=20000, $\eta$=0.001.\label{fig:Attack-on-Ladder}}
\end{figure}

In the rest of this section, we present some results about some attacks
on Ladder. Figure~\ref{fig:Attack} shows that Ladder is robust against
the boosting attack. Figure~\ref{fig:Attack-on-Ladder} shows some
brute-force enumeration attacks on Ladder. In these examples, to defend
against the attacks, Ladder uses less samples than necessary, for
the brute-force is not very efficient, since it does not make use
of all information available on the leaderboard.

\section{Sample complexity\label{sec:Sample-complexity}}

In this section, we present the sample complexity of the Ladder leaderboard.
We show that $\eta$ is \emph{not} only the display precision of the
leaderboard, but also the optimal value of $\eta$ \emph{is} the lowest
\emph{leaderboard error} possible. This optimal value is $\eta^{*}=\tilde{O}\left(\sqrt[3]{\frac{M}{n}}\right)$,
where $M$ is the number of accounts and $n$ is the number of samples
in the validation set.

Suppose that our data $(X,Y)$ lie in some space $\Omega\times\{0,1\}$.
They follow a distribution $\mathcal{D}$ on this space. Validation
set $S=\{(x_{1},y_{1}),\ldots,(x_{n},y_{n})\}$ are samples drawn
i.i.d. from this distribution. A classifier of this problem is represented
by the function $f:\Omega\rightarrow\{0,1\}$. The accuracy of this
classifier is defined as 
\[
R_{\mathcal{D}}(f):=\Pr(f(X)=Y).
\]
Its accuracy on the validation set is defined as
\[
R_{S}(f):=\frac{1}{n}\sum_{i=1}^{n}I(f(x_{i})=y_{i}),
\]
where $I(\cdot)$ is the indicator function.

$R_{S}(f)$ can be seen as an estimator of $R_{\mathcal{D}}(f)$,
whose error could be measured by the quantity $|R_{S}(f)-R_{\mathcal{D}}(f)|$.
Normally, this error should be small \cite{bousquet2002stability,bousquet2004introduction,devroye2013probabilistic,vapnik2013nature}.
However, due to the overfitting or hack by repeated and adaptive submissions,
this error could grow larger and larger. As this error grows, the
leaderboard is no longer a qualified index of the performance of participants.
The scores it displays no longer reflect the true accuracy of the
models, and the ranks it shows do not truly imply that one participant's
model is better than another's. This is why the traditional leaderboard
fails.

Since $R_{S}(f)$ is no longer a good estimator of $R_{\mathcal{D}}(f)$,
one may ask whether there exist other estimators. \cite{hardt2014preventing,steinke2014interactive}
show that no computationally efficient estimator can achieve error
$o(1)$ on more than $n^{2+o(1)}$ adaptively chosen functions in
a traditional leaderboard. Therefore, \cite{blum2015ladder} as well
as this article tries another approach: the Ladder leaderboard. 

In Ladder, the leaderboard only displays the best ever score that
an account has achieved $R_{t}:=\max_{1\le i\le t}R_{S}(f_{i})$,
where $f_{i}$ is the function which characterizes the $i$-th submission
associated with a given account. Thus, at the moment $t$, the error
of the score displayed on the leaderboard could be measured by the
quantity $|R_{t}-\max_{1\le i\le t}R_{\mathcal{D}}(f_{i})|$. Across
the time, the \emph{leaderboard error} of $R_{1},\ldots,R_{k}$ of
a single account is measured with
\[
\mathrm{lberr}(R_{1},\ldots,R_{k}):=\max_{1\le t\leq k}\left|R_{t}-\max_{1\le i\le t}R_{\mathcal{D}}(f_{i})\right|.
\]
A small leaderboard error means that the score displayed on the leaderboard
is close to the best score the account in question gets on the underlying
true distribution.

\cite{blum2015ladder} gives an upper bound to the leaderboard error.
However, it does not take into consideration the fact that a participant
may possess multiple accounts, in which case their reasoning breaks.
In this paper, we take that into consideration and our upper bound
is slightly tighter than theirs (logarithmic factor) when degenerated
to one-person-one-account case.

If a participant has multiple accounts, he can then switch among different
accounts to submit successive submissions. His submission thus does
not depend only on the history of the current account, but also on
the histories of \emph{other} accounts. Denote 
\[
\mathfrak{F}_{t}=\left\{ f_{i}^{m}:1\le i\le k_{m},1\le m\le M,\sum_{m=1}^{M}k_{m}=t\right\} 
\]
 as the submission history right after the moment $t$, where $f_{i}^{m}$
signifies using account $m$ to submit account $m$'s $i$-th submission,
and $k_{m}$ is the subtotal submissions associated with account $m$.
Denote 
\[
\mathfrak{R}_{t}=\left\{ R_{i}^{m}:1\le i\le k_{m},1\le m\le M,\sum_{m=1}^{M}k_{m}=t\right\} 
\]
 as the feedback (score) history associated with $\mathfrak{F}_{t}$. 
\begin{thm*}
Given a competition with a validation set of $n$ samples, where the
Ladder leaderboard employs a display precision of $\eta$ , and each
participant can possess at most $M$ accounts. For any set of $k$
(adaptively chosen) classifiers $\mathfrak{F}_{k}$ submitted by a
participant, his scores $\mathfrak{R}_{k}$ displayed on the leaderboard
satisfy 
\begin{equation}
\Pr\left\{ \max_{1\le t\leq k_{m},1\le m\le M}\left|R_{t}^{m}-\max_{1\le i\le t}R_{\mathcal{D}}(f_{i}^{m})\right|>\eta\right\} \le\exp\left(-\frac{\eta^{2}n}{2}+\left(\frac{M}{\eta}+1\right)\log2k+1\right).\label{eq:upper}
\end{equation}
In particular, for some $\eta=O\left(\sqrt[3]{\frac{M\log k}{n}}\right)$,
Ladder achieves with high probability: for any $m=1,\ldots,M$, 
\[
\mathrm{lberr}(R_{1}^{m},\ldots,R_{k_{m}}^{m})\le O\left(\sqrt[3]{\frac{M\log k}{n}}\right).
\]
\end{thm*}
Here, we successfully thrust the number of submissions $k$ into the
logarithmic factor. Thus, the leaderboard error is no longer sensitive
to the number of submissions. A participant can submit as many times
as he wishes (if he is able of course). However, we notice that the
number of accounts $M$ is still outside of the logarithmic factor,
which means that the leaderboard error grows quickly when $M$ grows.
Although it is only an upper bound, which is less persuasive than
a lower bound, we can provide a counter example to illustrate the
effect of multi-account. Consider the extreme case where the participant
submits each submission with a brand new account every time (i.e.,
$M=k$), the leaderboard error grows quickly with the submissions.
Indeed, Ladder shows no difference from the traditional leaderboard
in this case.

The theorem's aim is to prove a small leaderboard error. But why does
it matter? What is the relation to the robustness of a leaderboard?
The consequence of a small leaderboard error means that a participant's
score \emph{sticks to} his score on the ground truth. It is not likely
that he could climb up on the leaderboard either by overfitting or
by hack. If he marks a leap on the leaderboard, chances are that he
really improved his prediction model. 

Therefore, the theorem can be interpreted as: when the display precision
of ladder is set to the optimal value $\eta^{*}=O\left(\sqrt[3]{\frac{M\log k}{n}}\right)$,
a participant who gets a score $s$ (an integer multiple of $\eta^{*}$)
has large chance that his true score on the ground truth is within
$\left[s-\eta^{*},s+\eta^{*}\right]$. If two participants A and B
get the score $s_{A}$ and $s_{B}$ respectively, where $s_{A}-s_{B}>2\eta^{*}$,
chances are that A really outperforms B. Particularly, a hacker has
little chance to get a score higher than $\frac{1}{2}+\eta^{*}$,
since he learns nothing and his true score should be the same as the
random guess.

Naturally, we would like $\eta^{*}$ small. This depends on $n$.
We see that to achieve a small $\eta^{*}=\epsilon$, we will need
$O\left(\frac{M\log k}{\epsilon^{3}}\right)$ samples. 

\section{Proof of Theorem\label{sec:Proof-of-Theorem}}

From the state-of-art literature \cite{bousquet2002stability,bousquet2004introduction,devroye2013probabilistic,vapnik2013nature},
we already have
\begin{equation}
\Pr\left\{ \max_{1\le t\le k}\left|R_{S}(h_{t})-R_{\mathcal{D}}(h_{t})\right|>\varepsilon\right\} \le2k\exp(-2\varepsilon^{2}n),\label{eq:hoef}
\end{equation}
for a series of functions $h_{1},\ldots,h_{k}$ which are independent
of the validation set $S$. This inequality is quite close to our
destination. However, because of the sequential and adaptive nature
of our problem, $h_{t+1}$ is a function of $R_{S}(h_{1}),\ldots,R_{S}(h_{t})$,
which means that it is \emph{not} independent of $S$. Thus, we cannot
apply the above inequality directly. The technique employed in this
proof is to eliminate the dependence by enumerating all possible realizations. 
\begin{proof}
Suppose that the participant has an algorithm $\mathcal{A}$ to decide
the next function which characterizes the next submission, and which
account to submit with, in using the past history: $h_{t+1}=\mathcal{A}(\mathfrak{R}_{t}),$
where $h_{t+1}$ will become one member of the $f_{i}^{m}$ in $\mathfrak{F}_{t+1}$.
$\mathcal{A}$ can be either deterministic or random provided that
it does not depend on $S$. 

$h_{t+1}$ is dependent on $S$, however, 
\[
g:=\mathcal{A}\left\{ r_{i}^{m}:1\le i\le k_{m},1\le m\le M,\sum_{m=1}^{M}k_{m}=t\right\} 
\]
 is \emph{not}, where $r_{i}^{m}$ is one realization of $R_{i}^{m}$.
In consequence, we can apply (\ref{eq:hoef}) to $g$. The remaining
issue is to \emph{count} how many different $g$'s we could have within
$k$ submissions. 

To this end, we use a compression algorithm to encode every possible
$g$. First of all, to specify at which submission this $g$ is submitted,
we need $\left\lceil \log k\right\rceil $ bits. Then, each $g$ can
have history coming from $M$ different accounts. For each single
account, we calculate the bits needed. Since the history within an
account is a monotone increasing series, which should be multiples
of $1/\eta$ and be in interval $\left[0,1\right]$, it can only take
value in $\left\lceil 1/\eta\right\rceil $ numbers and jump at most
$\left\lfloor 1/\eta\right\rfloor $ steps. Now we calculate the number
of bits required to encode each jump. Here, we use a trick, which
allows us to get rid of the $n$ inside the logarithm in \cite{blum2015ladder}
\textendash{} we only code the place where there is a jump regardless
of jumping height. If the jump height is $1/\eta$, then we code this
place once. If the jump height is $s/\eta$, then we code it $s$
times. To code this place once, we need at most $\left\lceil \log k\right\rceil $
bits. Thus, the total bits demanded is
\[
\left\lceil \log k\right\rceil +M\times\left\lfloor \frac{1}{\eta}\right\rfloor \times\left\lceil \log k\right\rceil \le(\frac{M}{\eta}+1)\log2k.
\]

Then, we can apply (\ref{eq:hoef}) to the $g$'s in setting $\epsilon=\eta/2$:
\[
\Pr\left\{ \max_{g}\left|R_{S}(g)-R_{\mathcal{D}}(g)\right|>\frac{\eta}{2}\right\} \le2\times2^{(\frac{M}{\eta}+1)\log2k}\exp\left(-\frac{\eta^{2}n}{2}\right).
\]
The left side equals exactly 
\[
\Pr\left\{ \max_{h\in\mathfrak{F}_{k}}\left|R_{S}(h)-R_{\mathcal{D}}(h)\right|>\frac{\eta}{2}\right\} ,
\]
while the right side is bounded by
\[
\exp\left(-\frac{\eta^{2}n}{2}+\left(\frac{M}{\eta}+1\right)\log2k+1\right),
\]
which is exactly the right side of (\ref{eq:upper}). 

Conditioned on the event $\left\{ \max_{h\in\mathfrak{F}_{k}}\left|R_{S}(h)-R_{\mathcal{D}}(h)\right|\le\frac{\eta}{2}\right\} $,
we have
\[
\max_{1\le t\le k_{m},1\le m\le M}\left|\max_{1\le i\le t}R_{S}(f_{i}^{m})-\max_{1\le i\le t}R_{\mathcal{D}}(f_{i}^{m})\right|\le\frac{\eta}{2}.
\]
And since we have 
\[
\left|R_{t}^{m}-\max_{1\le i\le t}R_{S}(f_{i}^{m})\right|\le\frac{\eta}{2}
\]
 because of the rounding error, by the triangular inequality, we get
\[
\max_{1\le t\le k_{m},1\le m\le M}\left|R_{t}^{m}-\max_{1\le i\le t}R_{\mathcal{D}}(f_{i}^{m})\right|\le\eta,
\]

which is exactly what we want on the left side of (\ref{eq:upper}).
\end{proof}

\section{Discussion}

Ladder is vulnerable if the number of accounts each participant can
hold is unlimited. It is possible to launch a boosting attack on Ladder
leaderboard by using a brand-new account for each submission (Figure~(\ref{fig:Attack})).
Given limited number of accounts (e.g. one account for each participant),
Ladder is robust against the number of submissions. However, compared
with the quadratic sample complexity of general statistical / machine
learning tasks, the cubic sample complexity of the Ladder leaderboard
may still remain a bit too expensive. For a mere 0.01 leaderboard
error, the validation set has to have $10^{6}$ samples. This is not
possible in most competitions. Even if it is, we may still have questions
such as why not put these samples into the training dataset so as
to enable the participants to use more complex models. This makes
the loss of Ladder more than its gain. On the other hand, we do not
know whether this cubic root upper bound is tight. We have yet found
an efficient attack algorithm that could achieve this upper bound,
nor have we discovered a tight lower bound. That is to say, Ladder
may actually work better than we expected here. In practice, the competition
hosts employ as well other measures, such as limiting the number of
submissions per day, disqualifying participants secretly signing up
other accounts etc., to strengthen the accuracy of the leaderboard.
These measures could be combined with Ladder so as to provide a more
accurate leaderboard than the traditional one.

\bibliographystyle{unsrt}
\bibliography{leaderboard}

\end{document}